\documentclass[runningheads]{llncs}
\usepackage{graphicx}
\usepackage{hyperref}

\usepackage{todonotes}
\usepackage{amssymb}
\usepackage{tabularx}
\usepackage{subfigure}
\newcolumntype{L}{>{\raggedright\arraybackslash}X}

\makeatletter
\newcommand{\printfnsymbol}[1]{%
  \textsuperscript{\@fnsymbol{#1}}%
}
\makeatother

\begin{document}
\title{Hostility Detection in Hindi leveraging Pre-Trained Language Models\thanks{Shared Task in CONSTRAINT 2021}}
%
%
\author{Ojasv Kamal\thanks{Equal Contribution} \and Adarsh Kumar\printfnsymbol{2} \and Tejas Vaidhya}
\authorrunning{Kamal et al.}
%
\institute{Indian Institute of Technology, Kharagpur, West Bengal, India \\
\email{\{kamalojasv181, adarshkumar712.ak, iamtejasvaidhya\}@gmail.com}}
\maketitle              
%
\begin{abstract}
Hostile content on social platforms is ever increasing. This has led to the need for proper detection of hostile posts so that appropriate action can be taken to tackle them. Though a lot of work has been done recently in the English Language to solve the problem of hostile content online, similar works in Indian Languages are quite hard to find. This paper presents a transfer learning based approach to classify social media (i.e Twitter, Facebook, etc.) posts in Hindi Devanagari script as Hostile or Non-Hostile. Hostile posts are further analyzed to determine if they are Hateful, Fake, Defamation, and Offensive. This paper harnesses attention based pre-trained models fine-tuned on Hindi data with Hostile-Non hostile task as Auxiliary and fusing its features for further sub-tasks classification. Through this approach, we establish a robust and consistent model without any ensembling or complex pre-processing. We have presented the results from our approach in CONSTRAINT-2021 Shared Task\cite{patwa2021overview} on hostile post detection where our model performs extremely well with\textbf{ 3rd runner up} in terms of Weighted Fine-Grained F1 Score\footnote{refer Section \ref{Ex} for description of Weighted Fine-grained f1-score}.

\end{abstract}
\section{Introduction}
\label{Intro}
Social media is undoubtedly one of the greatest innovations of all time. From connecting with people across the globe to sharing of information and knowledge in a minuscule of a second, social media platforms have tremendously changed the way of our lives. This is accompanied by an ever-increasing usage of social media, cheaper smartphones, and the ease of internet access, which have further paved the way for the massive growth of social media. To put this into numbers, as per a recent report\footnote{\url{https://datareportal.com/reports/digital-2020-october-global-statshot}}, more than 4 billion people around the world now use social media each month, and an average of nearly 2 million new users are joining them every day. \\

While social media platforms have allowed us to connect with others and strengthen relationships in ways that were not possible before, sadly, they have also become the default forums for holding high-stakes conversations, blasting polarizing opinions, and making statements with little regard for those within the screenshot. The recent increase in online toxicity instances has given rise to the dire need for adequate and appropriate guidelines to prevent and curb such activities. The foremost task in neutralising them is hostile post detection. So far, many works have been carried out to address the issue in English \cite{Nobata2016AbusiveLD,waseem-etal-2017-understanding} and several other languages \cite{ref_arabic,ref_german}. Although Hindi is the third largest language in terms of speakers and has a significant presence on social media platforms, considerable research on hate speech or fake content is still quite hard to find. A survey of the literature suggests a few works related to hostile post detection in Hindi, such as \cite{JHA20202324,safi-samghabadi-etal-2020-aggression}; however, these works are either limited by inadequate number of samples, or restricted to a specific hostility domain.\\

A comprehensive approach for hostile language detection on hostile posts, written in Devanagari script, is presented in \cite{bhardwaj2020hostility}, where the authors have emphasized multi-dimensional hostility detection and have released the dataset as a shared task in Constraint-2021 Workshop. This paper presents a transfer learning based approach to detect Hostile content in Hindi leveraging Pre-trained models, with our experiments based on this dataset. The experiments are subdivided into two tasks, \textbf{Coarse Grained task}: Hostile vs. Non-Hostile Classification and \textbf{Fine Grained subtasks}: Sub-categorization of Hostile posts into fake, hate, defamation, and offensive.
\\

\noindent Our contribution comprises of improvements upon the baseline in the following ways:\\
\\
1. We fine-tuned transformer based pre-trained, Hindi Language Models for domain-specific contextual embeddings, which are further used in Classification Tasks.\\
2. We incorporate the fine-tuned hostile vs. non-hostile detection model as an auxiliary model, and fuse it with the features of specific subcategory models (pre-trained models) of hostility category, with further fine-tuning.\\

Apart from this, we have also presented a comparative analysis of various approaches we have experimented on, using the dataset. The code and trained models are available at this https url\footnote{\url{https://github.com/kamalojasv181/Hostility-Detection-in-Hindi-Posts.git}}

\section{Related Work}

In this section, we discuss some relevant work in NLP for Pre-Trained Model based Text Classification and Hostile Post Detection, particularly in the Indian Languages.

\subsection*{Pretrained-Language Models in Text Classification} Pre-trained transformers serve as general language understanding models that can be used in a wide variety of downstream NLP tasks. Several transformer-based language models such as GPT \cite{ref_Radford2018}, BERT \cite{devlin-etal-2019-bert}, RoBERTa \cite{DBLP:journals/corr/abs-1907-11692}, etc. have been proposed. Pre-trained contextualized vector representations of words, learned from vast amounts of text data have shown promising results in the task of text classification. Transfer learning from these models has proven to be particularly useful in tasks where there is a lack of undisputed labeled data and the inability of surface features to capture the subtle semantics in the text as in the case of hate speech \cite{ref_hate}. However, all these pre-trained models require large amounts of monolingual corpus to train on. Nonetheless, Indic-NLP \cite{ref_indic_Nlp} and Indic-Transformers \cite{jain2020indictransformers} have curated datasets, trained embeddings, and created benchmarks for classification in multiple Indian languages including hindi. \cite{Joshi_2020} presented a comparative study of various classification techniques for Hindi, where they have demonstrated the effectiveness of Pre-trained sentence embedding in classification tasks.     

\subsection*{Hostile Post Detection}Researchers have been studying hate speech on social media platforms such as Twitter \cite{ref_related_work}, Reddit \cite{ref_related_work_2}, and YouTube \cite{ref_related_work_4} in the past few years. Furthermore, researchers have recently focused on the bias derived from the hate speech training datasets \cite{ref_related_work_5}. Among other notable works on hostility detection, Davidson et al. \cite{Davidson} studied the hate speech detection for English. They argued that some words might reflect hate in one region; however, the same word can be used as a frequent slang term. For example, in English, the term ‘dog’ does not reveal any hate or offense, but in Hindi (ku\begin{small}\#\#\end{small}a) is commonly referred to as a derogatory term in Hindi. Considering the severity of the problem, some efforts have been made in Non-English languages as well \cite{ref_german,ref_arabic,ref_bangla,safi-samghabadi-etal-2020-aggression}. Bhardwaj et al. \cite{bhardwaj2020hostility} proposed a multi-dimensional hostility detection dataset in Hindi which we have focused on, in our experiments. Apart from this, there are also a few attempts at Hindi-English code-mixed hate speech \cite{ref_bohra}.

\section{Methodology}
\label{A}
In the following subsections, we briefly discuss the various methodologies used in our experiments. Each subsection describes an independent approach used for classification and sub-classification tasks. Our final approach is discussed in Section \ref{Aux}. 

\subsection{Single Model Multi-Label Classification}
\label{A_2}
{In this approach, we treat the problem as a Multi-label classification task. We use a single model with shared parameters for all classes to capture correlations amongst them. We fine tuned the pre-trained BERT transformer model to get contextualized embedding or representation by using attention mechanism. We experimented with three different versions of pre-trained BERT transformer blocks, namely Hindi BERT(a compressed form of BERT)\cite{HindiBERT}, Indic BERT(based on the ALBERT architecture)\cite{ref_indic_Nlp}, and a HindiBERTa model\cite{HindiBERTa}. The loss function used in this approach can be formulated mathematically as:

\[L(\hat{y},y) = - \sum_{j=1}^{c}y_{j}log\hat{y_j} + (1 - y_j)log(1-\hat{y_j})\]
\[J(W^{(1)}, b^{(1)}, ...) = 1/m \sum_{i = 1}^{m} L(\hat{y^{i}}, y^{(i)})\]

where, \textit{c} is total number of training examples 
and \textit{m} is number of different classes (i.e. non-hostile, fake, hate, defamation, offensive)
}

\subsection{Multi-Task Classification}
{In this approach, we considered the classification tasks as a Multi-task Classification problem. As described in Figure \ref{fig:model_b}, we use a shared BERT model and individual classifier layers, trained jointly with heuristic loss. This is done so as to capture correlations between tasks and subtasks in terms of contextualized embeddings from shared BERT model while maintaining independence in classification tasks. We experimented with Indic-BERT and HindiBERTa (we dropped the Hindi BERT model in this approach as the performance was poor compared to the other two models because of shallow architecture). The heuristic loss can be formulated mathematically as:

\[ L = l(x, y)  = { \{ l_1, ..., l_N\}}^T\]
where,  \[l_n = -w_n [y_n \cdot log\sigma(x_n) + (1- y_n) \cdot log(1- \sigma(x_n))]\]

\[L_{total} =  L_{(hostile/non-hostile)}+ \lambda \cdot 1/N \{L_{(hurt, defame, fake, offensive)}\}\]

if  post is Hostile \(\lambda = 0.5  \) (contributing to fine grain task ),  otherwise \(\lambda = 0\)
}

\subsection{Binary Classification}
{Unlike the previous two approaches, here we consider each classification task as an individual binary classification problem based on fine tuned contextualised embedding. We fine tuned the BERT transformer block and the classifier layer above it using the binary target labels for individual classes. Same as in Multi-task approach, we experimented this approach with Indic-BERT and HindiBERTa. Binary cross-entropy loss used in this approach can be mathematically formulated as follows:
\[L_{i}(\hat{y},y) = - \sum_{j=1}^{c}y_{j}log\hat{y_j} + (1 - y_j)log(1-\hat{y_j})\]

where, \textit{c} is total number of training examples and \textit{i} is number of independent models for each task
}

\subsection{Auxiliary Task Based Binary Sub-Classification}
\label{Aux}
{Similar to the previous approach, each classification task is considered as an individual binary classification problem. However, as an improvement over the previous approach, we treat the coarse-grained task as an Auxiliary task and then fuse its logits to each of the fine-grained subtasks. The motivation is that a hostile sub-class specific information shall be present in a post only if the post belongs to hostile class\cite{Kaushal_2020}. So, treating it as an Auxiliary task allow us to exploit additional hostile class-specific information from the logits of Auxiliary model. The loss function used in this case was same as described in Binary Classification. The model is described in Figure \ref{fig:model_a}.
\begin{figure}[ht]
    \centering
    \subfigure[]{%
        \includegraphics[width=0.4\linewidth]{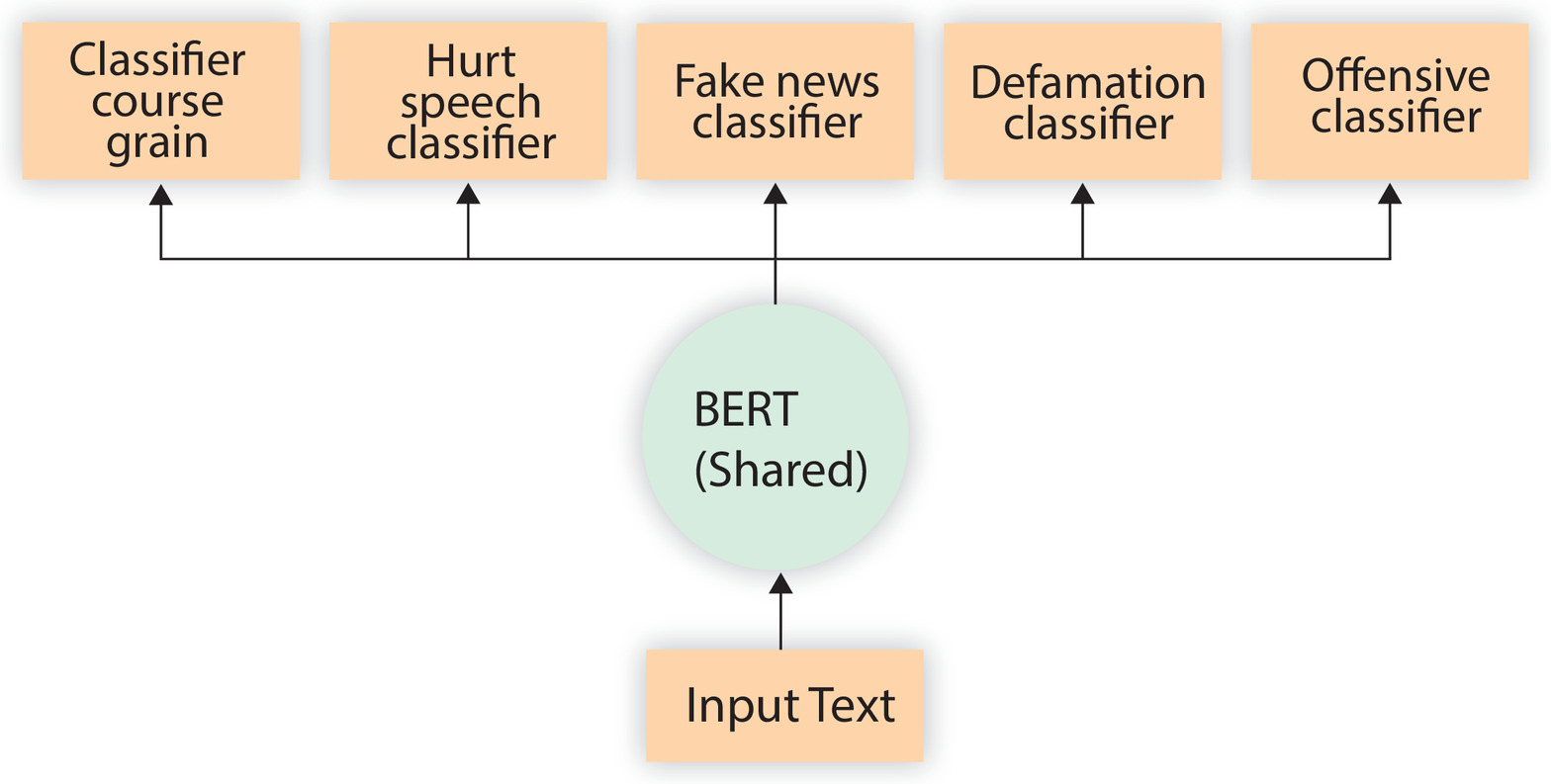}%
        \label{fig:model_b}%
        }%
    \hspace{5mm}%
    \subfigure[]{%
        \includegraphics[width=0.4\linewidth]{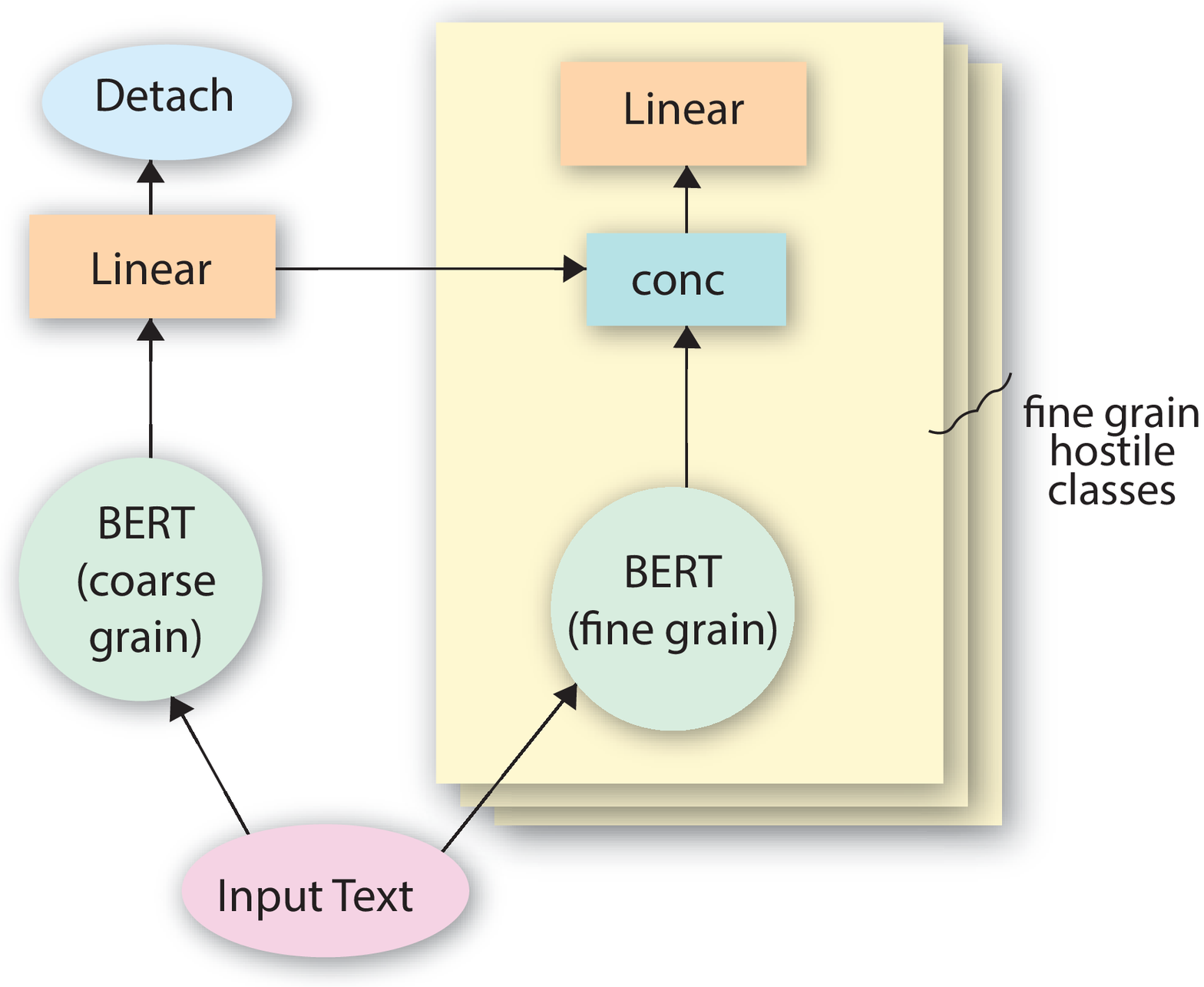}%
        \label{fig:model_a}%
        }%
    \caption{(a) Multi-Task Classification Model (b)Auxiliary Task Based Binary sub classification model.}
    \label{fig:model}
\end{figure}
}

\section{Experiment}

In this section, we first introduce the dataset used and then provide implementation details of our experiments in their respective subsections. 

\subsection{Dataset Description}

As already mentioned in Section \ref{Intro}, we evaluate our approach based on the dataset proposed in \cite{bhardwaj2020hostility}. As described in the dataset paper, the objective of the task is a classification of posts as Hostile and Non-Hostile and further Multi-label classification of Hostile posts into \emph{fake}, \emph{hate}, \emph{offensive}, and \emph{defame} classes. The dataset consists of 8192 online posts out of which 4358 samples belong to the non-hostile category, while the rest 3834 posts convey one or more hostile dimensions. There are 1638, 1132, 1071, and 810 posts for fake, hate, offensive, and defame classes in the annotated dataset, respectively. Same as in the paper \cite{bhardwaj2020hostility}, we split the dataset into 70:10:20 for train, validation, and test, by ensuring the uniform label distribution among the three sets, respectively.

\subsection{Pre-Processing}
Prior to training models, we perform the following pre-processing steps:
\begin{itemize}
    \item[$\bullet$] We remove all non-alphanumeric characters except full stop punctuation marks ( {$\vert$} , ? ) in Hindi, but we keep all stop words because our model trains the sequence of words in a text directly.
    \item[$\bullet$] We replace all user mentions and hashtags with a blank space.
    \item[$\bullet$] We skip emojis, emoticons, flags etc from the posts.
    \item[$\bullet$] We replace the URLs with the string `http`. 
    
\end{itemize}

\subsection{Experimental Setup}
\label{Ex}
All the experiments were performed using Pytorch \cite{ref_Pytorch} and HuggingFace \cite{ref_huggingface} Transformers library. As the implementation environment, we used Google Colaboratory tool which is a free research tool with a Tesla K80 GPU and 12GB RAM. Optimization was done using Adam \cite{adam} with a learning rate of 1e-5. As discussed earlier in section \ref{A}, in our experiments, we used pre-trained HindiBert\cite{HindiBERT}, IndicBert\cite{ref_indic_Nlp} and HindiBERTa \cite{HindiBERTa} Models available in HuggingFace library. Input sentences were tokenized using respective tokenizers for each model, with maximum sequence length restricted to 200 tokens. We trained each classifier model with a batch size of 16. In all the approaches, we used only the first token output provided by each model as input to classifier layer. Each classifier layer has 1 dropout layer with dropout of 0.3 and 1 fully connected layer. Each sub-classification task (fine grained task) was trained only on the hostile labeled examples, i.e. the posts that had at least one label of hostile class, so as to avoid extreme class-imbalance caused by including non-hostile examples. For the evaluation, we have used weighted f1 score \cite{scikit-learn} as a metric for measuring the performance in both the classification tasks. As suggested in the CONSTRAINT-2021 shared task\cite{patwa2021overview}, to measure the combined performance of 4 individual fine-grained sub-tasks together, we have used weighted fine-grained f1 score as the metric, where the weights for the scores of individual classes are the fraction of their positive examples.

\section{Results}
\begin{table}[bt]
\caption{\label{tab:table-name-2}Results obtained using various methods and models used. Here, \textbf{Baseline}: as described in the dataset paper \cite{bhardwaj2020hostility}, \textbf{MLC}: Multi Label Classification, \textbf{MTL}: Multitask Learning,  \textbf{BC}: Binary Classification and \textbf{AUX}: Auxiliary Model}
\begin{tabularx}{\textwidth}{llcccccc}
\hline
Method & Model               & Hostile           & Defamation        & Fake              &
Hate              & Offensive         & Weighted          \\
\hline
Baseline     & -          & 0.8422            & 0.3992            & 0.6869            & 0.4926           & 0.4198            & 0.542          \\
\hline

MLC    & Hindi-BERT          & 0.952           & 0.0               & 0.7528            & 0.4206            & 0.5274           & 0.4912           \\
       & Indic-BERT          & 0.9581           & 0.3787           & 0.7228          & 0.3094        & 0.5152           & 0.513             \\
       & HindiBERTa             & 0.9507            & 0.3239            & 0.7317            & 0.4120            & 0.4106           & 0.5122            \\
       \hline
MTL    & Indic-BERT          & 0.9284           & 0.0513           & 0.3296           & 0.0               & 0.0               & 0.1260           \\
       & HindiBERTa             & 0.9421           & 0.31              & 0.6647            & 0.2353            & 0.5545            & 0.4738            \\
       \hline
BC     & Hindi-BERT          & 0.9359            & 0.130             & 0.7164            & 0.47698           & 0.5388            & 0.5169            \\
       & Indic-BERT          & 0.9520           & 0.3030            & 0.757             & 0.4745            & 0.5446           & 0.5618            \\
       & HindiBERTa             & 0.9421           & 0.2707           & 0.6596           & 0.3175          & 0.6098            & 0.4960            \\
\hline
AUX    & \textbf{Indic-BERT} & \textbf{0.9583} & \textbf{0.42} & \textbf{0.7741} & \textbf{0.5725} & \textbf{0.6120} & \textbf{0.6250} \\
       & HindiBERTa             & 0.9486          & 0.3855          & 0.7612          & 0.5663          & 0.5933          & 0.6086\\
       \hline\\
\end{tabularx}
\end{table}

In this section, we discuss the results from the different approaches proposed in section \ref{A}. Table \ref{tab:table-name-2} summarizes the obtained results for different approaches, along with the baseline \cite{bhardwaj2020hostility}. Since hostile/non-hostile posts are real phenomenon, we did not perform oversampling and undersampling techniques to adjust class distribution and tried to supply the dataset as realistic as possible. This was done to avoid overfitting (in case of oversampling) and the loss of crucial data (in case of undersampling). As it's clear from Table \ref{tab:table-name-2}, our best model based on approach described in section \ref{Aux} with Indic-BERT model outperforms the baseline as well as other approaches in both the tasks, i.e. Coarse Grained Task of Hostile vs. Non-Hostile Classification and Fine Grained Task of Hostile Sub-Classification. Moreover, our best model stands as the \textbf{3\textsuperscript{rd} runner up} in terms of Weighted fine grained f1 score in the CONSTRAINT-2021 shared task on Hostile Post detection ( Results can be viewed \href{https://drive.google.com/file/d/1KYi4A\_QgmGRgEsxylCoLI2ddO872hMaf/view}{here}\footnote{Our team name is \textbf{Monolith}}). 

\section{Error Analysis}
\begin{table}[b!]
\caption{\label{tab:table-name} Misclassified Samples from the dataset}
\begin{tabularx}{\textwidth}{c}
\begin{minipage}{1.0\textwidth}
      \includegraphics[width=\textwidth]{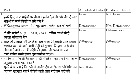}
    \end{minipage}
\end{tabularx}
\end{table}
Although we have received some interesting results, there are certain dimensions where our approach does not perform as expected. Through this section we try to better understand the obtained f1 scores through some general observations and some specific examples (refer Table \ref{tab:table-name}). Our model did perform comparatively better in fake dimension which implies the model was able to capture patterns in fake samples from dataset to a large extent. However, as can be seen in the example 1, the fake / non-fake classification of posts in certain cases largely context / knowledge based. Therefore, in absence of any external knowledge, the method is quite inefficient, particularly in those kind of samples which are under-represented in the dataset. Apart from this, we observe that the defamation scores are the lowest in general. This could be mainly attributed to the overall under-representation of the class in the dataset. Hence a more balanced dataset is critical to boost the defamation f1 score. \\

Another important observation to note is the existence of metaphorical data in the dataset, which implies meaning different from what semantic information is absent. For example, consider example 2 in the Table \ref{tab:table-name}. This tweet has been inspired by the Hindi idiom which means a person after committing every sin in the rule book looks to God for atonement and is used to refer to a hypocritical person indirectly. Such examples lead to mis-classification by models which are primarily based on contextualized embeddings training on simple datasets, as in our case. However, this could be eliminated if the models are pre-trained / fine-tuned on datasets which contain more such examples of metaphorical samples. From our manual inspection, we also observed that the dataset includes some examples, the labels of which are not even apparent to us. For instance, consider example 4. This example simply urges people to speak up and for some cause. Such type of sentence are quite often noticeable in hindi literature. It is impossible to conclude that it is an offensive post with the given data. However, the fact that it is correctly classified by our model reflects bias in the dataset with respect to certain kind of examples, against a generalization of the "Offensive" dimension. Apart from this, we also found some examples which, in our opinion are labeled incorrectly or are possibly ambiguous to be categorised in dimensions being considered. Example 5 means we do not want a favour we only ask for what we deserve which is labeled as defamation however according to us, it is ambiguous to classify it into any of the considered dimensions and largely dependent on the context. Similarly in example 6, someone is being referred as ku\begin{small}\#\#\end{small}e which means a dog, according to us it should be hate but is not labeled as hate.

\section{Conclusion and Future Work}

In this paper, we have presented a transfer learning based approach leveraging the pre-trained language models, for Multi-dimensional Hostile post detection. As the evaluation results indicate, our final approach outperforms baseline, by a significant margin in all dimensions. Furthermore, examining the results shows the ability of our model to detect some biases and ambiguities in the process of collecting or annotating dataset. \\

There is a lot of scope of improvement for fine Grained with few positive labels. Pre-training on relevant data (such as offensive or hate speech) is a promising direction. In case of Fake news detection, it is very difficult to verify the claim without the use of external knowledge. In future, we would like to extend the approach purposed in paper \cite{thorne2018fever}, by using processed-wikipedia knowledge it is possible to significantly improve fake news detection accuracy.

\section*{Acknowledgement}
We are very grateful for the invaluable suggestions given by Ayush Kaushal. We also thank the organizers of the Shared Task.
%
%
%
\bibliographystyle{splncs04}
\bibliography{References.bib}

\end{document}